\documentclass[lettersize,journal]{IEEEtran}
\usepackage{amsmath,amsfonts}
\usepackage{amssymb}
\usepackage{multirow}
\usepackage[ruled,vlined]{algorithm2e}
\DontPrintSemicolon
\usepackage{array}
\usepackage{newtxtext,newtxmath}
\usepackage[caption=false,font=normalsize,labelfont=rm,textfont=rm]{subfig}
\usepackage{textcomp}
\usepackage{stfloats}
\usepackage{url}
\usepackage{verbatim}
\usepackage{graphicx}
\usepackage{cite}
\usepackage{booktabs}
\usepackage{siunitx}
\begin{document}

\title{Dual-MPC Footstep Planning for Robust Quadruped Locomotion}

\author{Byeong-Il Ham$^{1}$, Hyun-Bin Kim$^{2}$, Jeonguk Kang$^{3}$, Keun Ha Choi$^{2}$, and Kyung-Soo Kim$^{2 *}$
}

\maketitle

\begin{abstract}
In this paper, we propose a footstep planning strategy based on model predictive control (MPC) that enables robust regulation of body orientation against undesired body rotations by optimizing footstep placement. Model-based locomotion approaches typically adopt heuristic methods or planning based on the linear inverted pendulum model. These methods account for linear velocity in footstep planning, while excluding angular velocity, which leads to angular momentum being handled exclusively via ground reaction force (GRF). Footstep planning based on MPC that takes angular velocity into account recasts the angular momentum control problem as a dual-input approach that coordinates GRFs and footstep placement, instead of optimizing GRFs alone, thereby improving tracking performance. A mutual-feedback loop couples the footstep planner and the GRF MPC, with each using the other’s solution to iteratively update footsteps and GRFs. The use of optimal solutions reduces body oscillation and enables extended stance and swing phases. The method is validated on a quadruped robot, demonstrating robust locomotion with reduced oscillations, longer stance and swing phases across various terrains.
\end{abstract}


\section{Introduction}
\IEEEPARstart{R}{esearch} on robust locomotion for legged robots has been actively conducted for several years~\cite{shin2022design, xu2023robust, grandia2023perceptive, risbourg2022real}. Notably, numerous studies have focused on achieving stable locomotion not only on flat terrain but also in complex and uncertain environments~\cite{choi2023learning, ding2024quadrupedal, ding2022robust}. Despite their effectiveness, locomotion strategies that rely on ground reaction forces (GRFs) suffer from inherent limitations. On low-friction surfaces, the magnitude of applicable GRFs is constrained to prevent slippage, which degrades stability~\cite{grandia2019feedback}. Abrupt transitions from high- to low-friction surfaces can induce slippage, thereby compromising stability during locomotion~\cite{su2023unified}. To address these limitations of GRF-based locomotion and enhance robustness, optimal footstep planning is required.

Exteroceptive sensor-based footstep planning focuses on enhancing safety by proactively recognizing uncertain terrain and planning footsteps in advance~\cite{fahmi2022vital}. In contrast, proprioceptive-based approaches focus on rapidly compensating for disturbances and model uncertainties~\cite{xin2021robust}. In this paper, we specifically focus on footstep planning based on the estimated state derived exclusively from proprioceptive sensors, aiming to reduce tracking errors and rapidly compensate for disturbances and model uncertainties.

Conventional footstep planning has commonly employed methods such as the Raibert heuristic~\cite{raibert1986legged}, zero moment point (ZMP)~\cite{vukobratovic1972stability}, and capture point (CP)~\cite{pratt2006capture}. These methods have limitations in compensating for angular momentum generated by uneven terrain or external disturbances. Such limitations hinder reliable performance in real-world environments. To address these issues, we present a footstep planning algorithm that takes angular velocity into account. Consequently, the robot achieves improved angular momentum regulation throughout locomotion. To realize this, we develop a footstep planner that incorporates system dynamics through model predictive control~(MPC).

By augmenting the state space, various studies have explored the use of a single MPC to simultaneously compute GRFs and footsteps\cite{grandia2023perceptive,kang2024external,kim2021design, ding2022orientation}. In these approaches, the time complexity is at least $\mathcal{O}(n^2)$, so the computation time increases quadratically or more with respect to the number of optimization variables. Real-world deployment requires all computations, including sensor processing, to be executed in real time on the onboard computer. Therefore, low time complexity is essential to ensure reliable operation under limited computational resources. To reduce computational load, footstep planning is performed using a dedicated MPC that solves only for footstep variables. The contributions of our study are as follows:
\begin{itemize}
    \item Extension of the angular momentum control problem from GRF-only to a dual-input, coupled formulation that coordinates GRFs and footstep placement, thereby improving tracking of angle and angular velocity.
    \item Iterative feedback between the GRF and footstep MPCs that simultaneously and progressively converges to optimal force and footstep solutions.
    \item Robust body orientation regulation that enables longer stance and swing phase scheduling.
\end{itemize}

\subsection{Related Works}
A fundamental method for deciding foot placement is the Raibert heuristic~\cite{raibert1986legged}, originally developed for dynamically stable hopping robots. This heuristic adjusts foot placement using velocity feedback to maintain balance and regulate body posture~\cite{yang2022fast}. Owing to its simplicity and computational efficiency, it remains widely adopted and is frequently integrated with other control strategies to enhance foot placement accuracy in complex environments. While effective for locomotion, it does not inherently optimize foot placement for energy efficiency or adaptability to varying terrains.

One widely adopted concept in legged locomotion is ZMP~\cite{vukobratovic1972stability,vukobratovic1969contribution}, defined as the point on the ground where the horizontal moment generated by the GRF is zero. Under this condition, no perturbation-induced tipping occurs. This method has been extensively used for legged robots to regulate center-of-mass (CoM) trajectories, maintaining balance~\cite{takenaka2009real,kajita2003biped,akbas2012zero}. However ZMP-based approaches exhibit limitations in dynamic scenarios, where maintaining strict zero moment conditions may restrict robustness. ZMP is difficult to compute and apply in dynamic gaits, where the assumptions required to define it no longer hold~\cite{gehring2014towards}.

In contrast, the CP concept~\cite{pratt2006capture}, commonly applied within the linear inverted pendulum model (LIPM), determines the optimal stepping location required to prevent falling based on the robot's momentum and CoM velocity~\cite{jeong2019robust,jeong2017biped, kim2025model}. By accounting for dynamic stability, this approach enables real-time footstep planning and allows robots to respond more effectively to external disturbances while maintaining balance. However, the LIPM is formulated without rotational inertia, making it impossible for the CP approach to incorporate angular momentum without extending the model.

Optimization-based techniques, including MPC, have been extensively applied to footstep planning, providing greater adaptability than purely heuristic-based methods. Convex MPC, formulated using a single rigid body model (SRBM), has been applied to simultaneously optimize foot placement and GRFs, improving adaptability in real-world environments~\cite{kim2021design, ding2022orientation}. Additionally, quadratic programming (QP) has been used to optimize foot placement using cost functions inspired by the Raibert heuristic~\cite{risbourg2022real}. A previous study has explored joint optimization of CoM trajectory and foot placement under ZMP constraints, ensuring stability and efficiency in locomotion~\cite{laurenzi2018quadrupedal}. Moreover, ZMP has been integrated with linear quadratic regulator and MPC to generate footstep while satisfying stability constraints~\cite{xin2021robust}. External force estimation has been incorporated into footstep optimization, as in the corrected capture point method, which adjusts step placement using real-time force measurements~\cite{kang2024external}. This approach enhances robustness by enabling the robot to adapt its step planning in response to external forces.

Machine learning-based approaches have also been explored for footstep planning and locomotion control. Artificial neural networks have been employed to predict foot placement and subsequently compute joint angles~\cite{tsounis2020deepgait}. The data-driven approach enables robots to generalize across different terrains and adapt to unstructured environments. Some studies have trained neural networks on diverse terrain conditions and deployed them for real-time locomotion using learned policies~\cite{cheng2024extreme,hoeller2024anymal}.

\subsection{Overview}
The remainder of this paper is organized as follows. In Section~\ref{sFootstepPlanner}, we present a model-based footstep planning approach formulated with MPC. Combining an MPC for footstep planning with an MPC for calculating the GRF is described in Section~\ref{sDoualModelPredictiveControl}. Validation and discussion are presented in Section~\ref{sValidation}. Finally, Section~\ref{sConclusion} concludes this study.

\section{Footstep Planner}
\label{sFootstepPlanner}
We present a footstep planning algorithm that utilizes approximated dynamics. By integrating the model and heuristic method within an MPC framework, the algorithm calculates the optimal footstep considering both angular and linear velocities.

\subsection{Approximated Dynamics}
We adopt the approximated SRBM for the robot model, which is expressed as follows:

\begin{equation}
    \ddot{\mathbf{p}} = \frac{1}{m}\sum{\mathbf{f}_i}-\mathbf{g}
    \label{model1}
\end{equation}

\begin{equation}
    \boldsymbol{\omega} \approx \mathbf{R}_z(\psi)\dot{\mathbf{\Theta}}
    \label{model2}
\end{equation}

\begin{equation}
    \mathbf{I}\dot{\boldsymbol{\omega}} \approx \sum{\mathbf{r}_i \times \mathbf{f}_i}
    \label{model3}
\end{equation}
\noindent
where~$\ddot{\mathbf{p}}$,~$m$,~$\mathbf{f}_i$,~$\mathbf{g}$,~$\boldsymbol{\omega}$,~$\boldsymbol{\dot{\omega}}$,~$\mathbf{\psi}$,~$\dot{\mathbf{\Theta}}$,~$\mathbf{I}$,~$\mathbf{r}_i$ are the three-dimensional body acceleration in world frame, body mass, i-th foot GRF in world frame, gravity vector, three-dimensional body angular velocity, angular acceleration, yaw angle, euler angle derivatives vector, $3 \times 3$~momentum of inertia matrix in world frame and three-dimensional i-th contact point with respect to the robot's CoM expressed in the world frame, respectively. Here, $i=1$~denotes the right-front leg, $i=2$~the left-front leg, $i=3$~the right-hind leg, and $i=4$~the left-hind leg. The conversion matrix from angular velocities to euler angle derivatives can be approximated as $\mathbf{R}_z(\psi)$, a counterclockwise rotation matrix of $\psi$ about the z-axis, under the assumption that the pitch angle is close to zero. When the angular velocities about the $x$ and $y$ axes are close to zero and the inertia matrix is nearly diagonal, the time derivative of $\mathbf{I}\boldsymbol{\omega}$ can be approximated as~(\ref{model3}), due to $\boldsymbol{\omega} \times \mathbf{I}\boldsymbol{\omega} \approx 0$~\cite{kim2019highly}.

\subsection{System Formulation}
In $\mathbb{R}^3$, the cross product can be expressed via the hat operator as
\begin{equation}
\label{crossProduct}
\mathbf{a} \times \mathbf{b} = \hat{\mathbf{a}}\mathbf{b} = -\hat{\mathbf{b}}\mathbf{a}\,, \quad \mathbf{a},\mathbf{b} \in \mathbb{R}^3
\end{equation}
where $\hat{\cdot}$ denotes the skew-symmetric matrix representation of a vector. Substituting $\mathbf{r}_i=\mathbf{p}_{b,i}-\mathbf{p}$ into~(\ref{model3}),where $\mathbf{p}_{b,i}$ denotes the placement of the $i$-th foot, the body position $\mathbf{p}$, and applying the property in~(\ref{crossProduct}) yields
\begin{equation}
    \sum\mathbf{r}_i \times \mathbf{f}_i = \sum (\mathbf{p}_{b,i}-\mathbf{p}) \times \mathbf{f}_i = -\sum\mathbf{\hat{f}}_i\mathbf{p}_{b,i} - \mathbf{\hat{p}}\sum\mathbf{f}_i \,.
    \label{extModel3}
\end{equation}
In~(\ref{extModel3}), the body position and GRFs are assumed to remain constant. The GRFs are obtained from the GRF-computing MPC, using the solution at the first horizon corresponding to the swing leg transition phase. More details on the GRF-computing MPC are provided in Section~\ref{ssGroundReactionForceComputation}.

Given a state $\mathbf{x} = [\mathbf{p} \; \dot{\mathbf{p}} \; \Theta \; \boldsymbol{\omega} \;1]^\top$, $\dot{\mathbf{p}}$ denotes the body velocity in world frame, $\Theta$ the Euler angles, and $1$ is dummy state, the SRBM can be represented by the following state-space equation:
\begin{equation}
    \frac{d}{dt}\mathbf{x}= \mathbf{A}(\mathbf{f},\psi)\mathbf{x}+\mathbf{B}(\mathbf{f},\psi)\mathbf{p}_b \,.
    \label{srbmdynamics}
\end{equation}
The dynamics matrix $\mathbf{A}(\mathbf{f},\psi)$ and input matrix $\mathbf{B}(\mathbf{f},\psi)$ are defined as
\begin{equation}
    \begin{aligned}
    \mathbf{A}(\mathbf{f},\psi) = \begin{bmatrix}
        \mathbf{0} & \mathbf{I} & \mathbf{0} & \mathbf{0} & \mathbf{0}_{3\times1} \\
        \mathbf{0} & \mathbf{0} & \mathbf{0} & \mathbf{0} & \frac{1}{m} \sum \mathbf{f}_i-\mathbf{g} \\
        \mathbf{0} & \mathbf{0} & \mathbf{0} & \mathbf{R}_z(\psi)^{\top} & \mathbf{0}_{3\times1} \\
        \mathbf{0} & \mathbf{0} & \mathbf{0} & \mathbf{0} & -\mathbf{I}^{-1}\hat{\mathbf{p}} \sum \mathbf{f}_i \\
        \mathbf{0}_{1\times3} & \mathbf{0}_{1\times3} & \mathbf{0}_{1\times3} & \mathbf{0}_{1\times3} & 0
    \end{bmatrix}\!
    \end{aligned}
\end{equation}

\begin{equation}
    \mathbf{B}(\mathbf{f},\psi) = 
    -\begin{bmatrix}
        \mathbf{0} & \mathbf{0} & \mathbf{0} & \mathbf{0} \\
        \mathbf{0} & \mathbf{0} & \mathbf{0} & \mathbf{0} \\
        \mathbf{0} & \mathbf{0} & \mathbf{0} & \mathbf{0} \\
        \mathbf{I}^{-1}\hat{\mathbf{f}}_1 & \mathbf{I}^{-1}\hat{\mathbf{f}}_2 & \mathbf{I}^{-1}\hat{\mathbf{f}}_3 & \mathbf{I}^{-1}\hat{\mathbf{f}}_4
    \end{bmatrix}\!.
\end{equation}
The zero matrices in rows 1 to 4 of the dynamics matrix, excluding the last column, are 3 by 3 matrices, while the remaining row is 1 by 3 matrix. In the input matrix, all zero matrices are 3 by 3. GRFs and yaw angle are assumed to remain constant over $N$~prediction horizons, allowing the system to be treated time-invariant. By applying zero-order hold, the system is represented in discrete time.

\subsection{Cost Function}
An MPC cost function for determining footstep is defined as follows:

\begin{equation}
    l_k(\mathbf{x},\mathbf{p}_b) = {\|\mathbf{x}^d-\mathbf{x}\|}_{\mathbf{Q}_k} + {\|\mathbf{p}_b\|}_{\mathbf{R}_k}
    \label{costFunction}
\end{equation}
where $\cdot^d$ denotes desired value, $\cdot_k$ is $k$-th horizon, $\mathbf{Q}_k \in \mathbb{R}^{13 \times 13}$ and $\mathbf{R}_k \in \mathbb{R}^{12 \times 12}$ are weight matrices. A norm is defined as $\|\mathbf{x}\|_{\mathbf{M}}=\mathbf{x}^{\top}\mathbf{Mx}$. In the SRBM (\ref{model1}), the foot placement does not affect the body acceleration. Therefore, a unique solution can not be guaranteed. Furthermore, ${\|\mathbf{p}_b\|}_{\mathbf{R}_k}$ varies with the body position, which can lead to variations in tracking performance.

To address these issues, a modified cost function is adopted by incorporating the heuristic method $\mathbf{p}^d_b$.

\begin{figure}[t!]
    \centering
        \includegraphics[width=1.0\columnwidth]{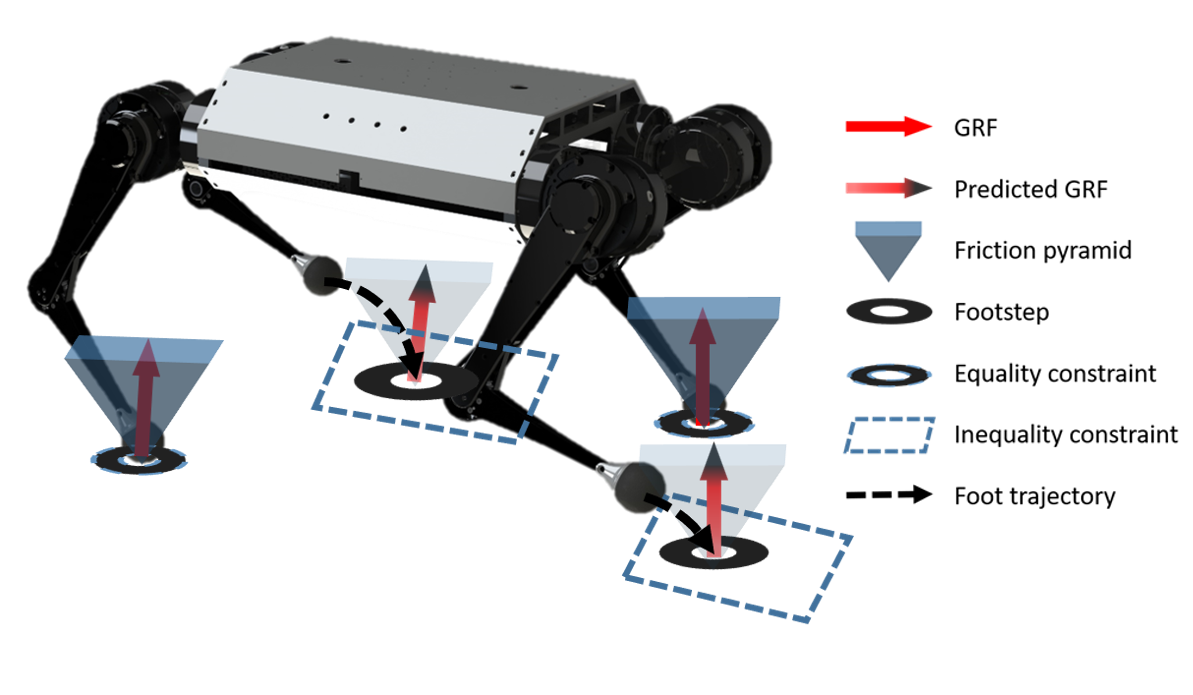}
        \caption{
        Illustration of the proposed method. The GRF and predicted GRF correspond to the solutions of the GRF MPC. The friction pyramid represents the GRF inequality constraint. The footstep denotes the swing-leg solution obtained from the footstep MPC, where the equality constraint is the current footstep of the stance leg, and the inequality constraint corresponds to the footstep inequality constraint.
        }    
    \label{titleFig}
\end{figure}

\begin{equation}
    l_k(\mathbf{x},\mathbf{p}_b^d,\mathbf{p}_b) = {\|\mathbf{x}^d-\mathbf{x}\|}_{\mathbf{Q}_k} + {\|\mathbf{p}^d_b-\mathbf{p}_b\|}_{\mathbf{R}_k} \, .
    \label{costFunctionModified}
\end{equation}
Integrating the heuristic method into the cost function enables foot placement to account for both angular and linear velocities, refining the optimal footstep toward a solution that considers angular velocity in the vicinity of the heuristic method.

\begin{equation}
    \mathbf{p}^d_{b} = \mathbf{p}_{hip} + \frac{t_{s}}{2}\dot{\mathbf{p}}+\mathbf{k}({\dot{\mathbf{p}}^d-\dot{\mathbf{p}}})+\frac{{\omega}^d_z}{2}\sqrt{\frac{{p}_z}{\|\mathbf{g}\|}}\dot{\mathbf{p}}^{\perp}
    \label{heuristicFoot}
\end{equation}
here $\mathbf{p}_{hip,i}$ denotes the position of the $i$-th leg’s hip, $t_{s}$ is the scheduled swing phase time, $\|\mathbf{g}\|$ is gravity constant and $\mathbf{k}$ is the diagonal gain matrix. For any $\mathbf{a} \in \mathbb{R}^3$, the perpendicular operator is defined as $\mathbf{a}^\perp = -\mathbf{e}_z \times \mathbf{a}$, where $\mathbf{e}_z$ is the unit vector along the $z$-axis.

\subsection{Constraint}

The footstep must lie within the kinematic workspace of the swing leg. Additionally, to ensure robustness against impulse disturbances, the solution is constrained to remain close to the CP. Unlike the cost function defined in the world frame, CP is defined in the body frame. Using Z-Y-X Euler angles, the rotation matrix from the body frame to the world frame can be expressed as follows:

\begin{equation}
    \mathbf{R(\Theta)} = \mathbf{R}_z(\psi)\mathbf{R}_y(\theta)\mathbf{R}_x(\phi)
\end{equation}
where $\theta$ and $\phi$ denote the pitch and roll angles, respectively, and $\mathbf{R}_k(\gamma) \in SO(3)$ represents a counterclockwise rotation of $\gamma$ about the $k$-axis. Through this rotation matrix, the transformation from the world frame to the body frame can be derived. Here, we will use the prescript ${}^b\cdot$ to indicate that it is expressed in the body frame:
\begin{equation}
    {}^b\dot{\mathbf{p}} = \mathbf{R^{\top}(\Theta)}\dot{\mathbf{p}}
\end{equation}
\begin{equation}
    {}^b\mathbf{p}_b = \mathbf{R^{\top}(\Theta)}(\mathbf{p}_b-\mathbf{p}) \,.
    \label{world2bodyposition}
\end{equation}

The constraint ensuring proximity to the CP is given by
\begin{equation}
    -\Delta{}^b\mathbf{p}_{max} \leq {}^b\mathbf{p}_{b} - {}^b\dot{\mathbf{p}}\sqrt{\frac{{p}_z}{\|\mathbf{g}\|}} \leq -\Delta{}^b\mathbf{p}_{min}\,.
    \label{inequalconst}
\end{equation}
The parameters $\Delta{}^b\mathbf{p}_{max}$ and $\Delta{}^b\mathbf{p}_{min}$ are computed from the body's maximum velocity and the swing phase time. Generalizing (\ref{inequalconst}), it can be formulated as follows:
\begin{equation}
    \mathbf{A}_q{}^b\mathbf{p}_b \leq \mathbf{b}_q
    \label{inequalconstgenbody}
\end{equation}
here $\mathbf{A}_q$ and $\mathbf{b}_q$ denote the inequality constraint matrix and vector, respectively. By substituting (\ref{world2bodyposition}) into (\ref{inequalconstgenbody}), the generalized inequality constraint in the world frame is obtained as:
\begin{equation}
    \mathbf{A}_q\mathbf{R^{\top}(\Theta)}\mathbf{p}_b \leq \mathbf{b}_q + \mathbf{A}_q\mathbf{R^{\top}(\Theta)p} \,.
    \label{worldinequalconstraints}
\end{equation}

The foot placement of the stance leg must remain fixed at the contact point unless slip occurs. Therefore, an equality constraint is imposed to maintain the fixed footstep throughout the stance phase.

\subsection{Footstep MPC Formulation}

MPC can be formulated as the following optimization problem:

\begin{equation}
\begin{aligned}
\min_{\mathbf{x},\mathbf{u}} \quad & l_N(\mathbf{x}_N,\mathbf{u}_{N}^d,\mathbf{u}_{N}) +  \sum_{k=0}^{N-1} {l_k(\mathbf{x}_k,\mathbf{u}_{k}^d,\mathbf{u}_{k})} \\
    \text{s.t.}\quad
    & \mathbf{x}_{k+1} = \mathbf{A}_d\mathbf{x}_k+\mathbf{B}_d\mathbf{u}_{k} , \;\;\;(\text{system dynamics}) \\
    & \mathbf{E}_k\,\mathbf{u}_k=\mathbf{c}_k \;, \qquad\qquad\:\;(\text{stance phase}) \\
    & \mathbf{G}_k\,\mathbf{u}_k \le \mathbf{h}_k \;, \qquad\qquad\:(\text{swing phase}),\; (\ref{worldinequalconstraints}) \\
    & \mathbf{u}\equiv\mathbf{p}_{b} \;, \; \mathbf{u}^d\equiv (\ref{heuristicFoot}) \; .
\end{aligned}
\label{optimizationProblem}
\end{equation}

This optimization problem is defined over $N$~prediction horizons. The system dynamics is given by the discrete-time form of (\ref{srbmdynamics}), with $\mathbf{u}_k$ representing $\mathbf{p}_b$ at the $k$-th step. $\mathbf{E}$, $\mathbf{c}$, $\mathbf{G}$, and $\mathbf{h}$ denote generic constraint matrices and vectors. The equality constraint is applied to the stance leg, while the inequality constraint is applied to the swing leg. In the cost function (\ref{costFunctionModified}), $\mathbf{Q}_k \gg \mathbf{R}_k$ is set so that the state error is minimized as much as possible, while the footstep remains close to the heuristic method.

The state space over $N$ horizons can be formulated as follows:
\begin{equation}
    \mathbf{X} = \mathbf{A}_{qp}\mathbf{x}_0+\mathbf{B}_{qp}\mathbf{U}
\end{equation}
where $\mathbf{X} = \begin{bmatrix} \mathbf{x}_1\; \mathbf{x}_2\; \cdots\; \mathbf{x}_N \end{bmatrix}^{\top}$, $\mathbf{U} = \begin{bmatrix} \mathbf{u}_0\; \mathbf{u}_1\; \cdots\; \mathbf{u}_{N-1} \end{bmatrix}^{\top}$, and $\mathbf{A}_{qp}~\in~\mathbb{R}^{13N \times 13}$ and $\mathbf{B}_{qp}~\in~\mathbb{R}^{13N \times 12N}$ are obtained from $\mathbf{A}_d$ and $\mathbf{B}_d$, respectively. The optimization problem in (\ref{optimizationProblem}) can be expressed in QP form as follows:
\begin{equation}
    \sum_{k=0}^{N-1} l_k(\mathbf{x}_{k}, \mathbf{u}_{k}) = \frac{1}{2}\mathbf{U}^\top\mathbf{PU} + \mathbf{U}^\top\mathbf{q} + \mathbf{C}
    \label{qpProblem}
\end{equation}
\begin{equation}
    \mathbf{P} = 2(\mathbf{B}^{\top}_{qp}\mathbf{Q}\mathbf{B}_{qp} + \mathbf{R})
    \label{matP}
\end{equation}
\begin{equation}
    \mathbf{q} = 2(\mathbf{B}^{\top}_{qp}\mathbf{Q}(\mathbf{A}_{qp}\mathbf{x}_0-\mathbf{X}^{d})-\mathbf{R}\mathbf{U}^d)
\end{equation}
here $\mathbf{Q} \in \mathbb{R}^{13N \times 13N}$ and $\mathbf{R} \in \mathbb{R}^{12N \times 12N}$ are diagonal matrices with positive definite properties, composed of $\mathbf{Q}_k$ and $\mathbf{R}_k$, respectively. $\mathbf{C}$ is a constant matrix independent of $\mathbf{U}$. If~(\ref{matP}) is positive semi-definite, the QP guarantees a global minimum solution~\cite{boyd2004convex}. A proof is provided in the Appendix~A.

\begin{figure*}[t!]
    \centering
        \includegraphics[width=1.92\columnwidth]{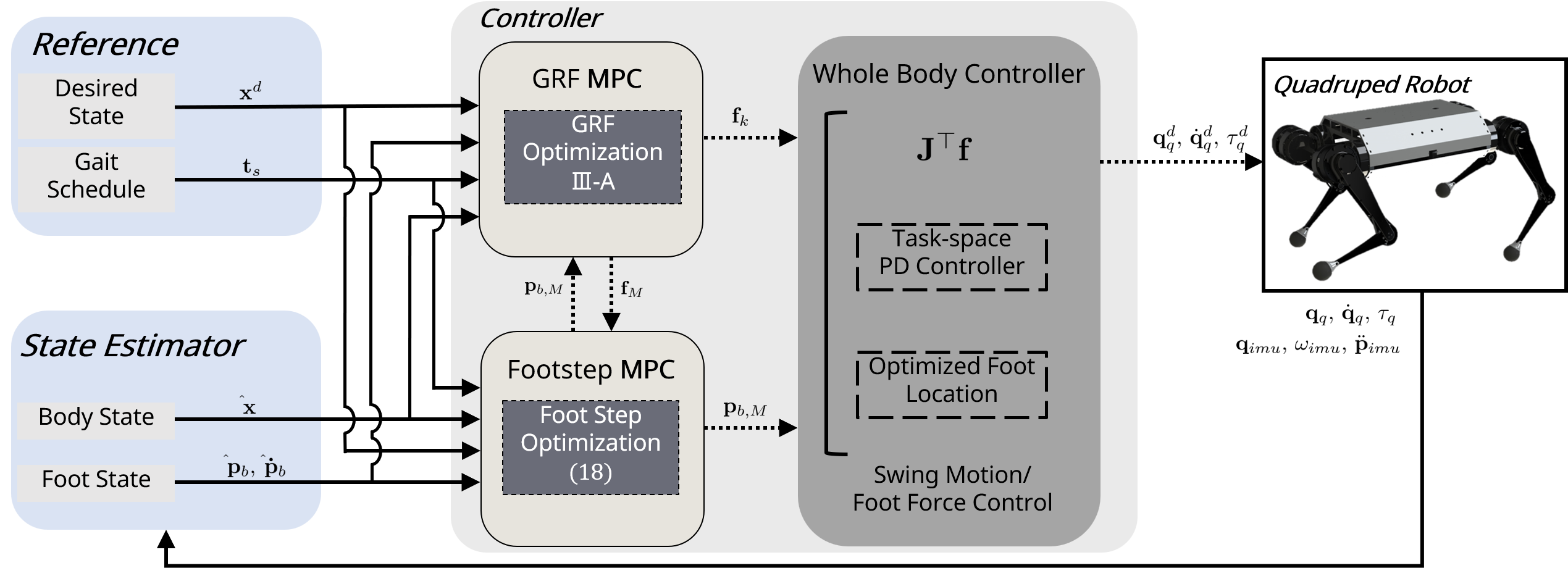}
        \caption{
        Overall locomotion architecture.
        }    
    \label{sysFrameworkFig}
\end{figure*}
\section{Dual model predictive control}
\label{sDoualModelPredictiveControl}

We propose a Dual MPC (DMPC) framework, which incorporates a feedback structure between two MPCs. One MPC is used for footstep planning, as described in Section~\ref{sFootstepPlanner}, and the other for computing the GRF. This section presents the GRF computation method and the integration of the two MPCs.

\subsection{Ground Reaction Force Computation}
\label{ssGroundReactionForceComputation}

The GRF can be computed using an MPC in a manner similar to the MPC introduced in Section~\ref{sFootstepPlanner}. By replacing the foot placement with the GRF in (\ref{srbmdynamics}), the dynamics can be written as:
\begin{equation}
    \frac{d}{dt}\mathbf{x} = \mathbf{A}(\psi)\mathbf{x} + \mathbf{B}(\mathbf{p}_b,\psi)\mathbf{f} \,.
    \label{srbmdynamicsforce}
\end{equation}
$\mathbf{p}_b$ is assumed to remain constant over $N$~horizons. The cost function in (\ref{costFunction}) is reformulated by replacing $\mathbf{p}_b$ with $\mathbf{f}$. Similar to the footstep planner, the constraints are defined separately for the stance and swing phases. Since the swing leg is not in contact with the ground, the GRF must be zero. Therefore, an equality constraint~$\mathbf{f} = \mathbf{0}$ is imposed during the swing phase. In the stance phase, the GRF is constrained such that no slip occurs at the contact point. Accordingly, the GRF in the stance phase is constrained to remain within the friction pyramid. Using the friction coefficient $\mu$, this constraint can be expressed as:
\begin{equation}
    |f_x| \leq \mu f_z \,\,, |f_y| \leq \mu f_z \,, f_z \geq 0\,.
    \label{frictionCoefficient}
\end{equation}
The force components along the x- and y-axes are constrained to remain within $\mu$ times the vertical force component $f_z$.

This MPC can likewise be formulated as an optimization problem over $N$ horizons, analogous to the footstep planner. Thus, it can be reformulated as a QP using the same method, and the optimal GRF can be obtained. The GRF is obtained using the solution at the first horizon. Further details are provided in~\cite{di2018dynamic}.

\subsection{Combination of Two Model Predictive Control}

The procedure for hierarchically combining the two MPCs is presented in Algorithm~\ref{algorithmofDMPC}. It details the mutual-feedback loop in which the footstep and GRF MPCs exchange solutions. For $N$ horizons, the $M$-th horizon, obtained from the gait scheduler, corresponds to the first horizon in which the phase transition occurs. Both MPCs use horizons of equal length. The GRF MPC optimizes over $\{k, k+1, \dots, k+N-1\}$, whereas the footstep MPC optimizes over the fixed shifted window $\{M, M+1, \dots, M+N-1\}$. The current GRF is taken from the first step of the GRF-QP solution, and the solution at the $M$-th horizon is used as the GRF when the swing leg makes ground contact. The swing leg contact point is obtained from the first step of the footstep-QP solution and is also applied to the system dynamics. The above process is visualized in Fig.~\ref{titleFig}, where the GRF denotes $\mathbf{u}_{f,k}$, the predicted GRF denotes $\mathbf{u}_{f,M}$, and the footstep denotes $\mathbf{u}_{p,M}$. Through iterative feedback, the GRF and footstep MPCs progressively converge to optimal solutions.

\begin{algorithm}[t!]
\caption{Dual MPC: Hierarchical integration of footstep and GRF optimization}
\label{algorithmofDMPC}

\While{locomotion}{
  \textbf{get} $M , \mathbf{x}^d$\;
  \For{$k \gets 0$ \KwTo $M$}{
    \textbf{get} $\mathbf{x}_0$\;
    \textbf{formulate GRF-QP} $\gets \mathbf{x}_0, \mathbf{x}^d$ \;
    \textbf{solve GRF-QP}  $\to \mathbf{U}_f$\;
    \textbf{update}  $\mathbf{f} \gets \mathbf{u}_{f,k}$\;
    \textbf{update dynamics}  (\ref{srbmdynamics}) $\gets \mathbf{u}_{f,M}$\;
    \textbf{compute} (\ref{heuristicFoot}) $\to \mathbf{p}^d_{b}$ \;
    \textbf{formulate footstep-QP}  (\ref{qpProblem}) $\gets \mathbf{x}_0, \mathbf{x}^d, \mathbf{p}^d_{b}$ \;
    \textbf{solve footstep-QP}  $\to \mathbf{U}_p$\;
    \textbf{update}  $\mathbf{p}_b \gets \mathbf{u}_{p,M}$\;
    \textbf{update dynamics}  (\ref{srbmdynamicsforce}) $\gets \mathbf{p}_b$\;    
  }
}
\end{algorithm}

\subsection{System Framework}

The locomotion architecture of the quadruped robot with the proposed method is shown in Fig.~\ref{sysFrameworkFig}. The architecture is divided into four main components: reference, state estimator, controller, and robot. The desired states generated using a joystick and the predefined swing/stance time vector $\mathbf{t}_s$ are provided as inputs to the MPCs. In addition, the estimated state together with the estimated foot position and velocity are also provided as inputs to the MPCs. The DMPC returns $\mathbf{f}$ and $\mathbf{p}_b$ as defined in Algorithm~\ref{algorithmofDMPC}. The low-level controller is provided with the DMPC outputs and the DMPC inputs together with the Jacobian matrix $\mathbf{J}$. Using these, the whole-body controller (WBC) computes the desired joint position $\mathbf{q}_q^d$, velocity $\dot{\mathbf{q}}_q^d$, and torque, which are then applied through the joint PD controller. The output of the low-level controller, desired joint torque $\mathbf{\tau}_q^d$ which generated by the PD controller, is transmitted to the quadruped robot. The robot then sends the joint states back to both the low-level controller and the state estimator, while the quaternion, gyroscope, and accelerometer measurements from the Inertial Measurement Unit (IMU) are provided as inputs to the state estimator. The orientation is computed using the quaternion from the IMU, while the robot’s body and foot positions/velocities are estimated via a linear Kalman filter. The state estimator and WBC run at 500 $\mathrm{Hz}$, the motor PD controller at 40 $\mathrm{kHz}$, and the DMPC at 20 $\mathrm{Hz}$.

\section{Experimental}
\label{sValidation}

The proposed footstep planning algorithm was experimentally validated on the Unitree GO1 robot. The proposed algorithm was executed on an Intel NUC with an Intel i7-1165G7 CPU running Ubuntu 20.04. The GO1 and the computation PC communicated via UDP. We adopted the QP solver qpSWIFT~\cite{pandala2019qpswift}, and variable definitions were implemented through the Eigen3 library. For performance evaluation, a baseline combining heuristic footstep planning with GRF MPC was used for comparison against the proposed approach. In both methods, the prediction horizon was set to 10, corresponding to a swing phase of 250$\mathrm{ms}$.

\begin{figure}[t!]
    \centering
    \subfloat[\label{Exp1SetupFigure}]{
        \includegraphics[width=0.47\linewidth]{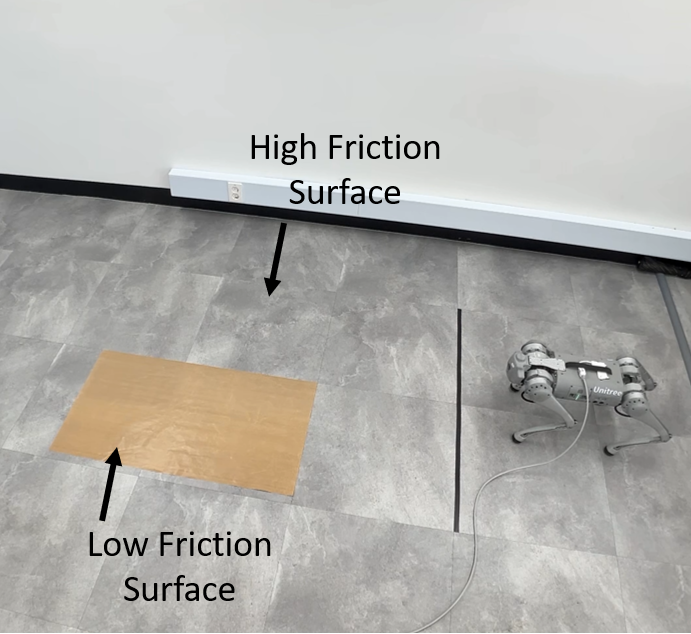}
    }%
    \subfloat[\label{Exp2SetupFigure}]{
        \includegraphics[width=0.52\linewidth]{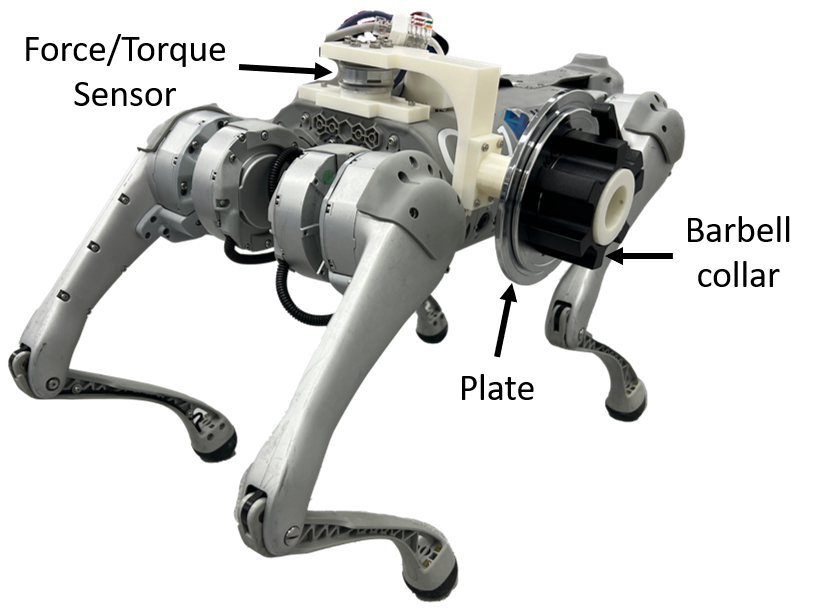}
    }%
    \caption{Experimental setup. (a)~Asymmetric friction terrain composed of high and low friction surfaces. The left side is the low-friction surface, while the right side is the high-friction surface. (b)~The F/T sensor is mounted at a location offset along the x-axis from the CoM.}
    \label{ExpSetupFigure}
\end{figure}
\begin{figure*}[t!]
    \centering
    \subfloat[\label{Exp1ControlVelFigure}]{
        \includegraphics[width=0.32\linewidth]{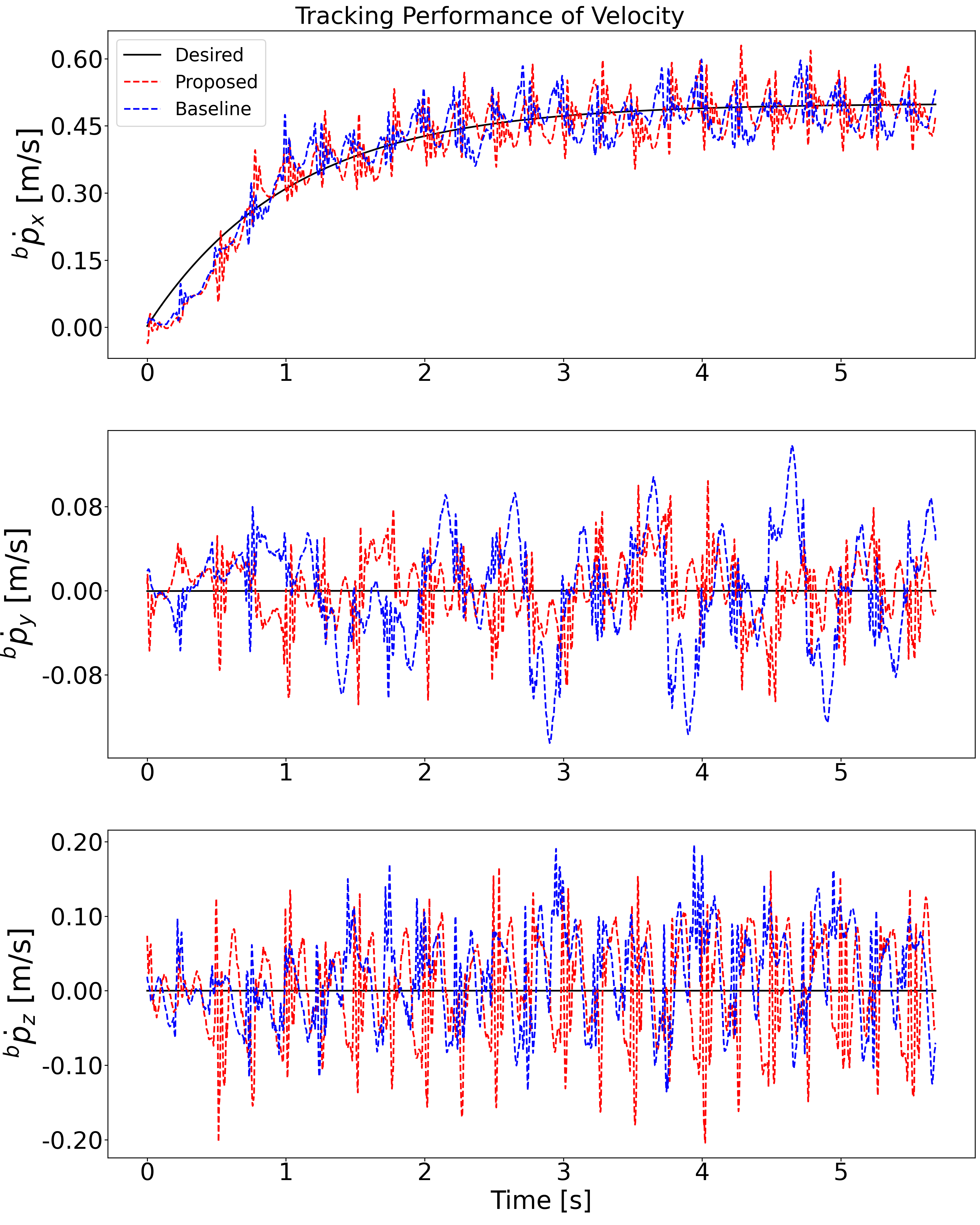}
    }%
    \subfloat[\label{Exp1ControlRPYFigure}]{
        \includegraphics[width=0.32\linewidth]{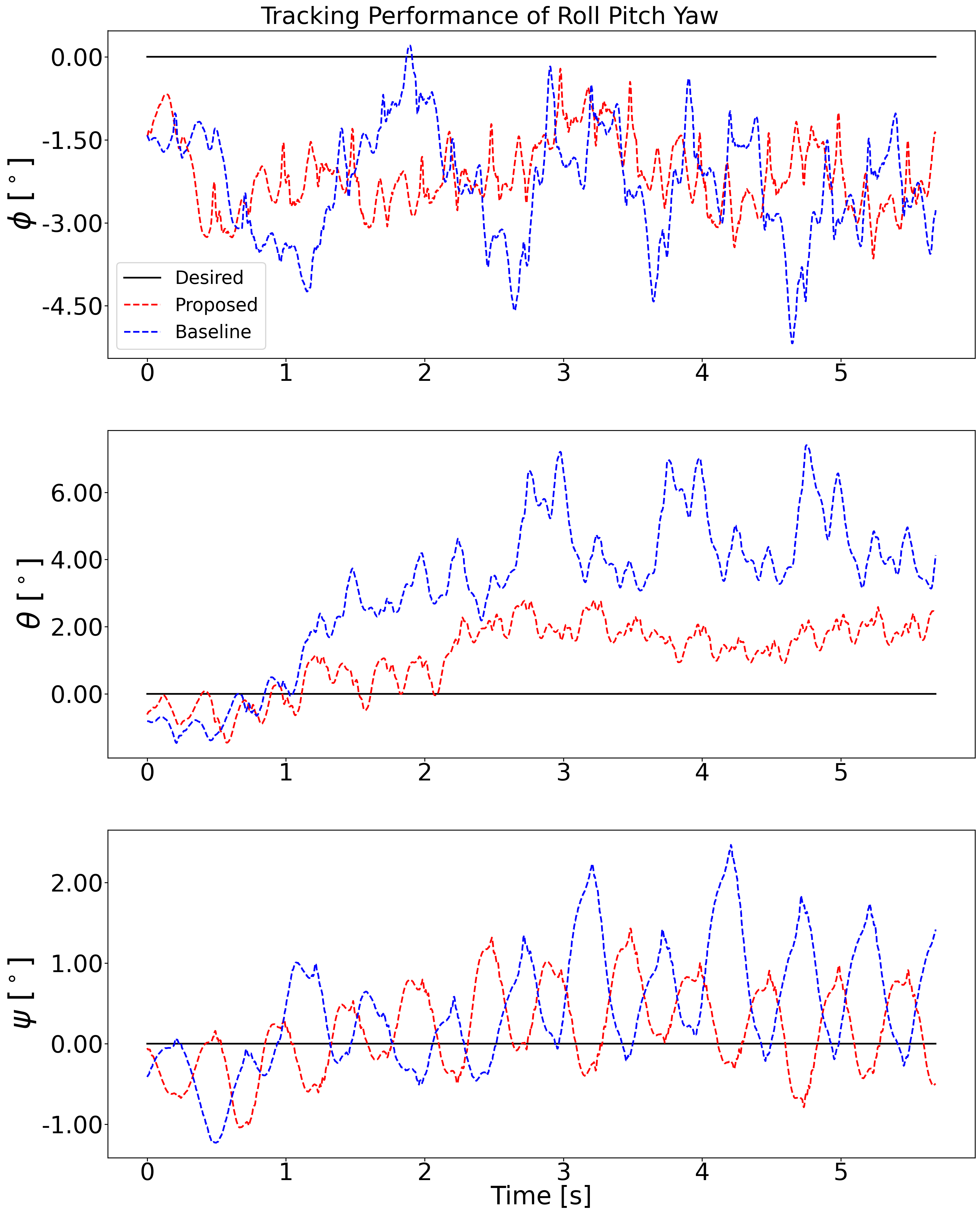}
    }%
    \subfloat[\label{Exp1ControlAngFigure}]{
        \includegraphics[width=0.32\linewidth]{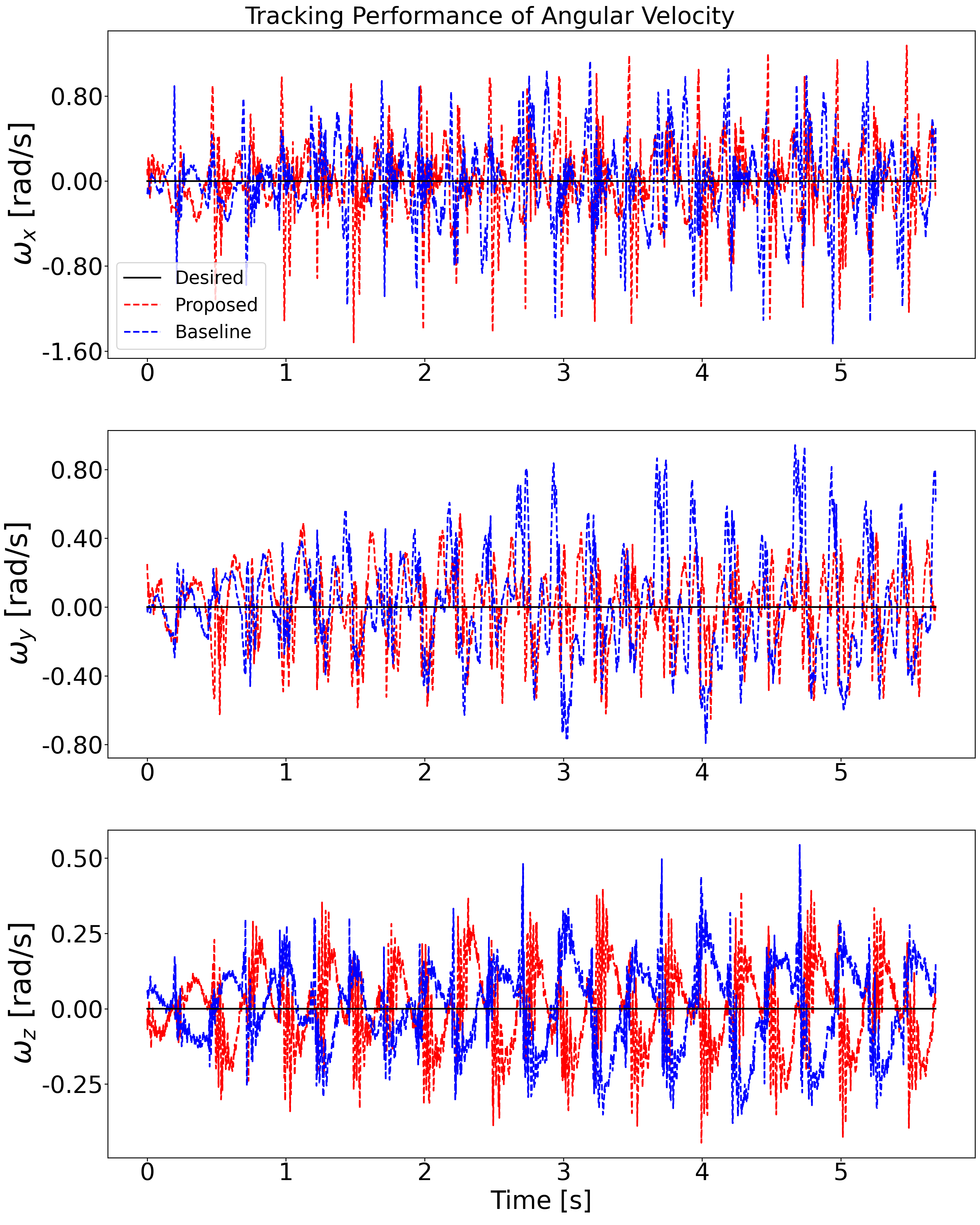}
    }%
    \caption{Experimental results of locomotion on asymmetric friction terrain, showing desired values and states. (a)~Velocity in body frame. (b)~Roll, pitch, and yaw. (c)~Angular velocity.}
    \label{Exp1ControlFigure}
\end{figure*}

\begin{figure}[t!]
    \centering
         \includegraphics[width=0.99\columnwidth]{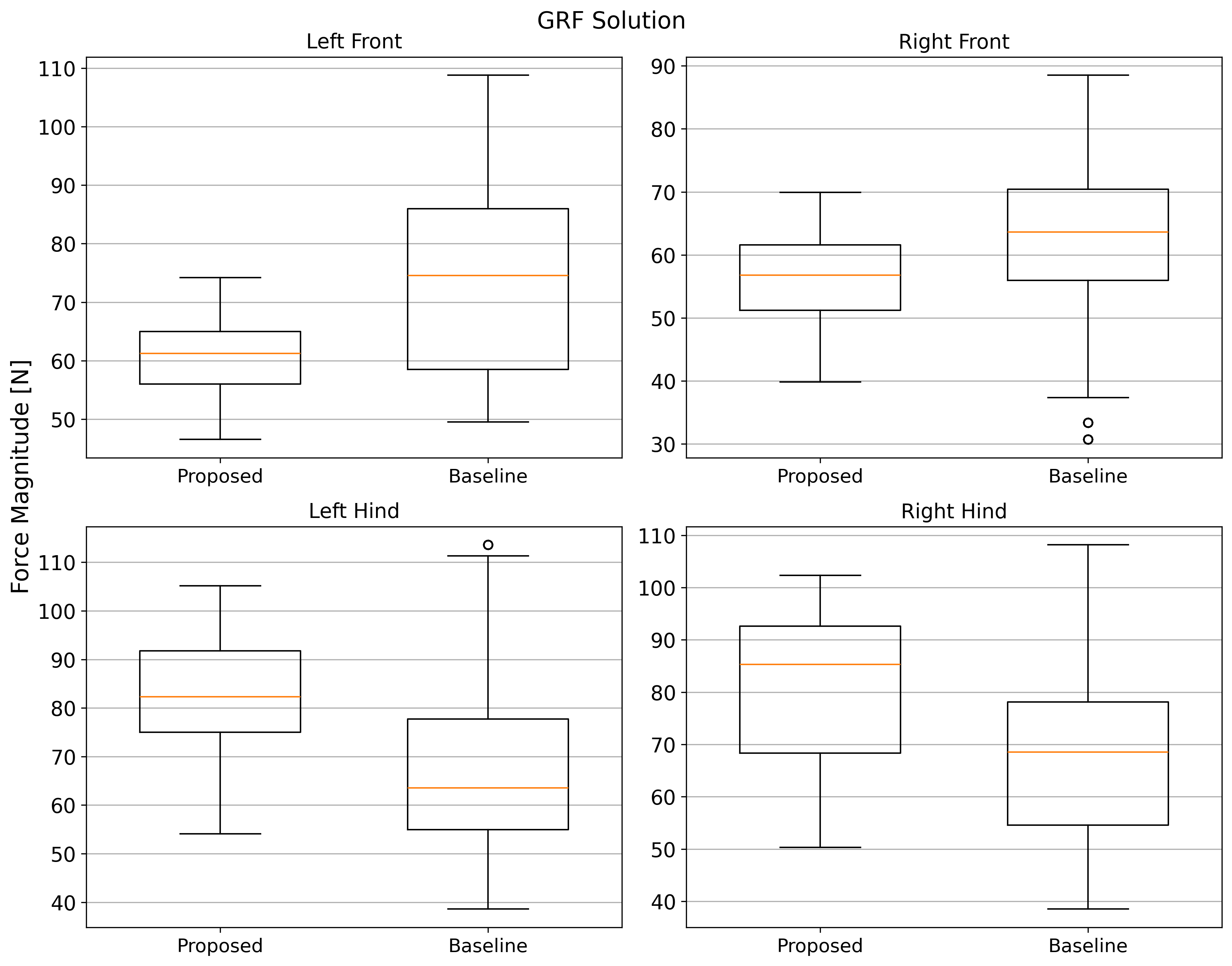}
        \caption{
        Boxplots of GRFs for each leg during locomotion, comparing the proposed method with the baseline.
        }    
    \label{Exp1ForceFigure}
\end{figure}

\begin{figure}[t!]
    \centering
        \includegraphics[width=1\columnwidth]{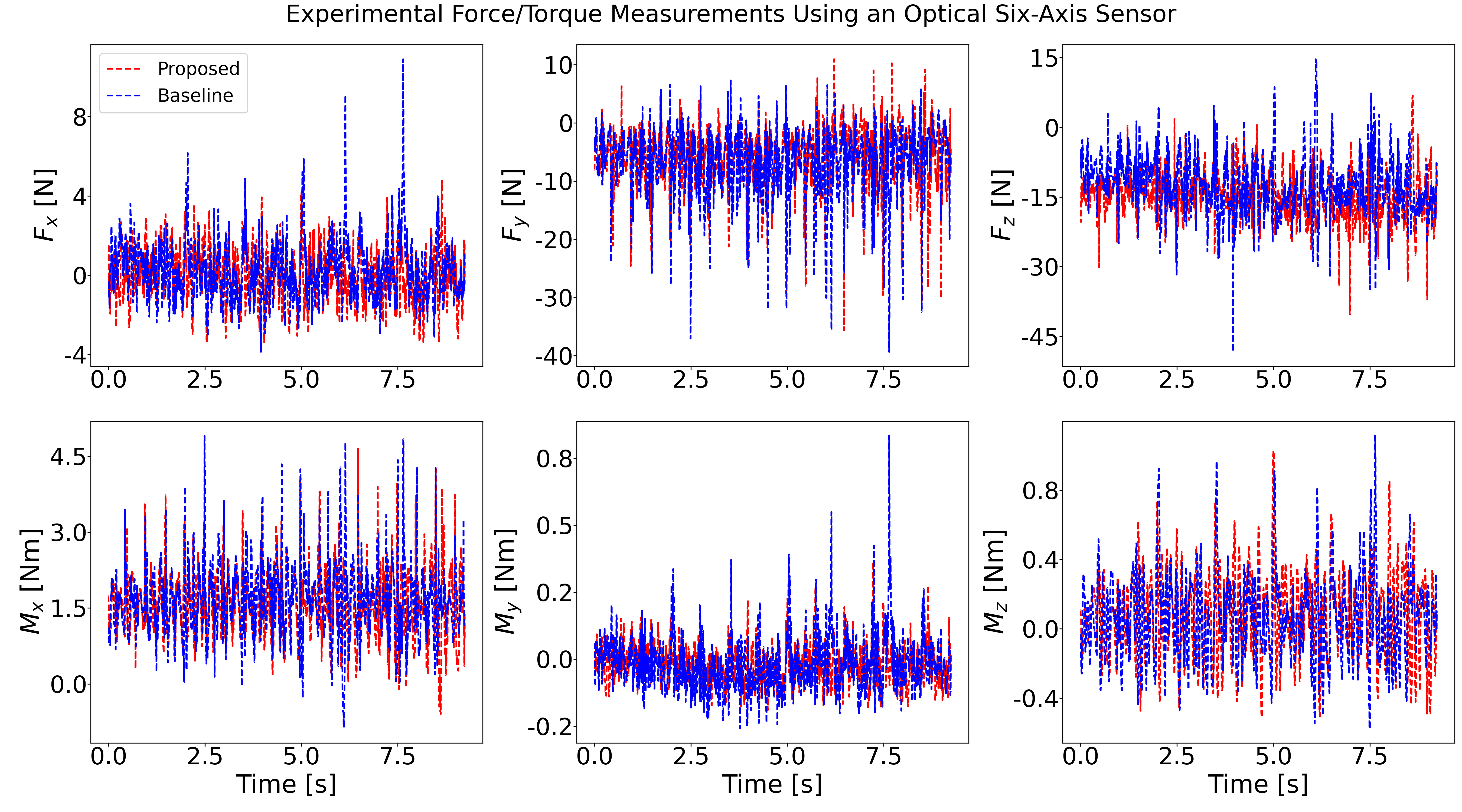}
        \caption{
        Measured forces and torques in the body frame obtained from the F/T sensor. The red dashed lines represent proposed experiment, while the blue dashed lines indicate baseline experiment.
        }
    \label{FTResultFigure}
\end{figure}

\subsection{Locomotion on Asymmetric Friction Terrain}
\label{Exp1}

To evaluate tracking performance, unexpected angular momentum was introduced by intentionally inducing slip on one side feet during locomotion. This was done by setting GRF-MPC $\mu$ in (\ref{frictionCoefficient}) higher than the friction coefficient of the low-friction surface, thereby causing the 2nd and 4th feet to slip, as illustrated in Fig.~\ref{Exp1SetupFigure}. The desired velocity was set to 0.5~$\mathrm{m/s}$ along the x-axis of the body frame.

Control performance was obtained as shown in Fig.~\ref{Exp1ControlFigure} and the mean squared error (MSE) is presented in Table~\ref{Exp1MSE}. Compared to the baseline, the MSEs of the proposed method are 135\%, -39\%, 2\%, -21\%, -84\%, -57\%, -6\%, -53\%, and -13\%, respectively, indicating a reduction in the tracking errors of $\mathbf{\Theta}$ and $\boldsymbol{\omega}$. Although the error of the proposed method increased in the x- and z-axis velocities, the velocity MSE for both methods remains lower than that of the other states.

Fig.~\ref{Exp1ForceFigure} shows the boxplots of the GRFs for the stance legs during locomotion, where the GRF magnitude is defined as the Euclidean norm of the force components. For the proposed method, the mean/standard deviation of the GRF magnitudes for each leg are 56.13/6.67, 60.78/6.54, 80.37/15.30, and 80.68/13.39~$\mathrm{N}$, whereas the baseline values are 61.99/12.23, 75.11/15.04, 68.79/18.92, and 68.83/18.60~$\mathrm{N}$. The sums of the mean values are 277.96~$\mathrm{N}$ and 274.72~$\mathrm{N}$, respectively, corresponding to a marginal difference of 1.18\%. However, in the baseline, the mean GRFs of the front and hind legs are 137.1~$\mathrm{N}$ and 137.62~$\mathrm{N}$, differing by only 0.34\%, whereas in the proposed method, they are 116.91~$\mathrm{N}$ and 161.05~$\mathrm{N}$, indicating that the front-leg GRF is 27.1\% smaller than that of the hind legs. Defining the force ratio as $\frac{|f_x|}{f_z}$, the per-leg means for the proposed method are 0.125, 0.114, 0.094, and 0.084, whereas those for the baseline are 0.134, 0.120, 0.103, and 0.120, indicating lower force ratios for the proposed method across all legs.

\subsection{Locomotion with Wrench Disturbance}

\begin{figure*}[t!]
    \centering
    \subfloat[\label{Exp2ErrorVelFigure}]{
        \includegraphics[width=0.32\linewidth]{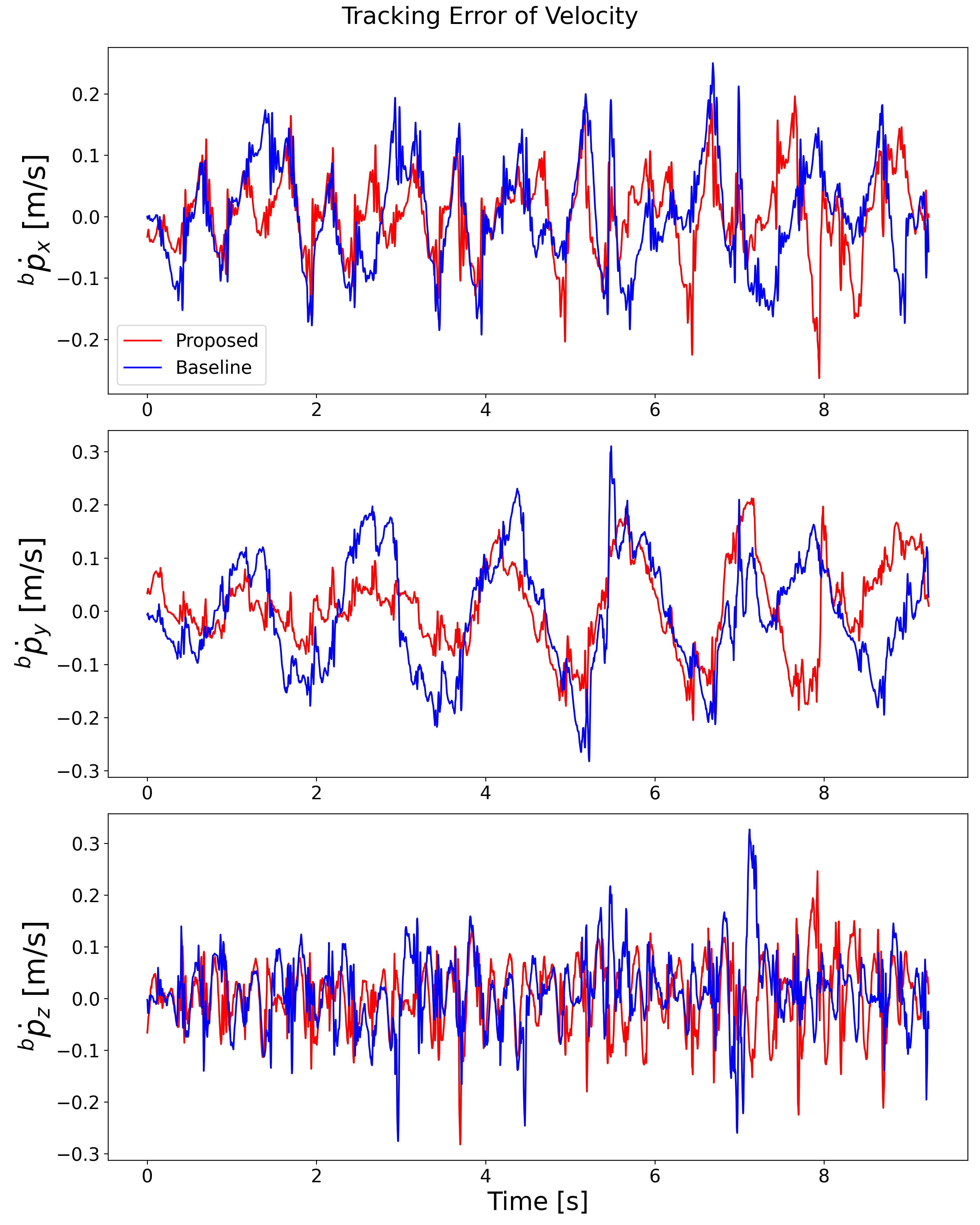}
    }%
    \subfloat[\label{Exp2ErrorRPYFigure}]{
        \includegraphics[width=0.32\linewidth]{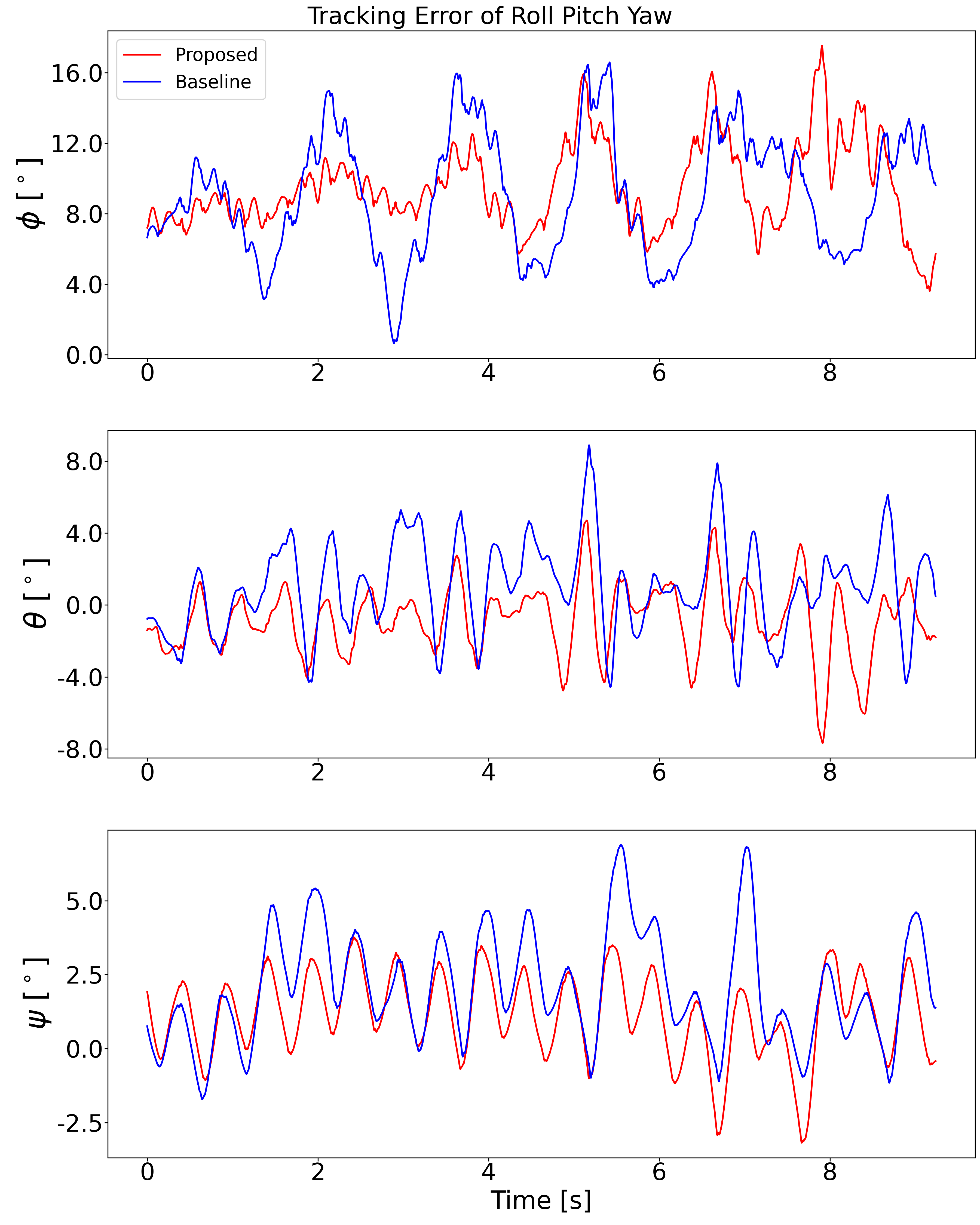}
    }%
    \subfloat[\label{Exp2ErrorAngFigure}]{
        \includegraphics[width=0.32\linewidth]{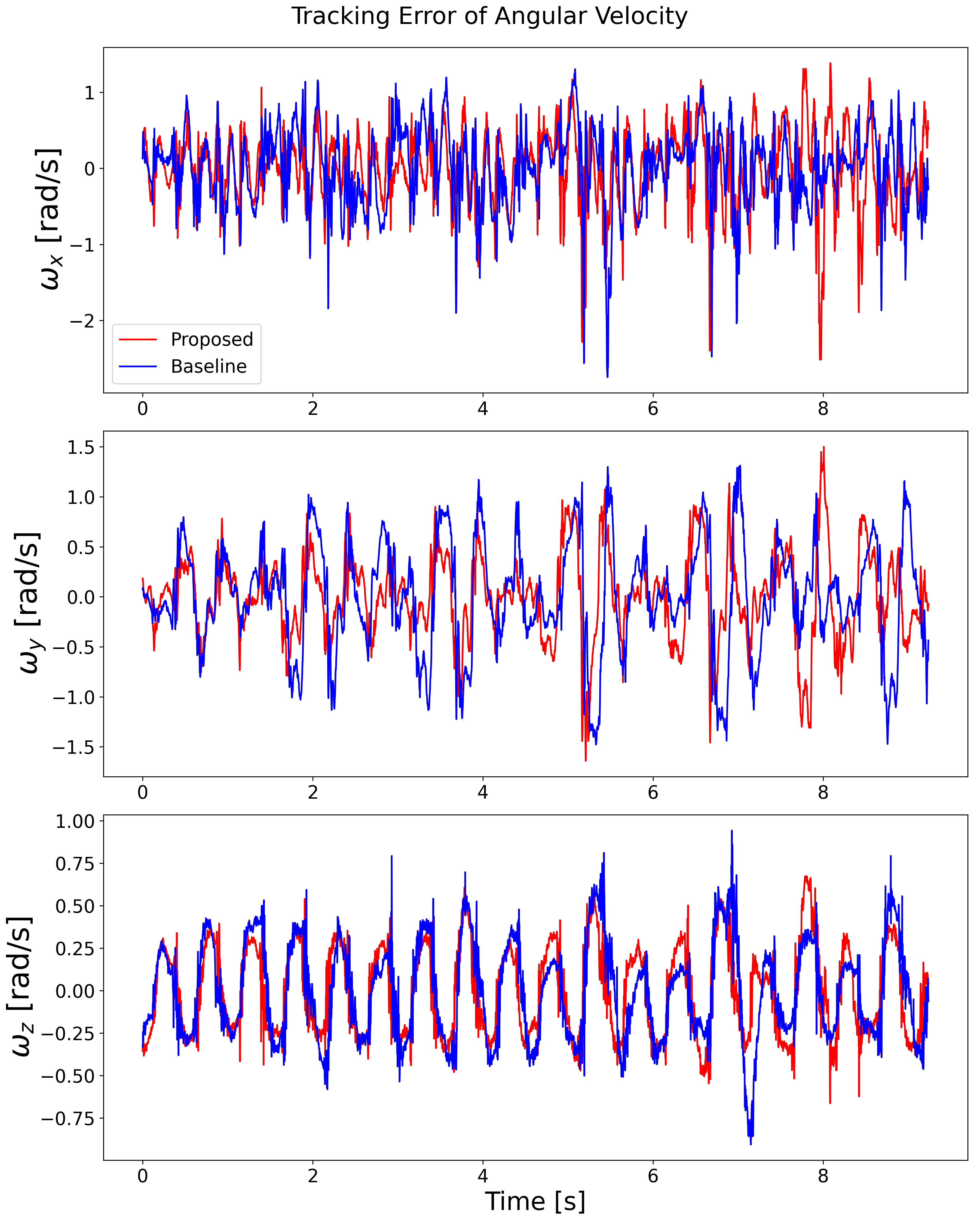}
    }%
    \caption{Experimental results of locomotion under wrench disturbance, illustrating tracking errors. (a)~Velocity in body frame. (b)~Roll, pitch, and yaw. (c)~Angular velocity.}
    \label{Exp2ErrorFigure}
\end{figure*}

An optical six-axis force/torque (F/T) sensor~\cite{kim2025compact,kim2024compact} was mounted on the robot body. The wrench disturbance was generated using plates, and a barbell collar, and measured by the F/T sensor Fig.~\ref{Exp2SetupFigure}. The F/T sensor was calibrated with the beam attached, so it measured only the forces and torques applied by the plates and the barbell collar to the robot body. One 0.5 $\mathrm{kg}$ plate, one 0.25 $\mathrm{kg}$ plate, and one 0.16 $\mathrm{kg}$ collar were used in this experiment. The desired velocity was set to 0.3~$\mathrm{m/s}$ along the x-axis of the body frame.

\begin{table}[!t]
\centering
\caption{MSE on asymmetric-friction terrain}
\resizebox{\columnwidth}{!} {
\begin{tabular}{cccccccccc}
\hline
                         & ${}^{b}\dot{p}_x$   & ${}^{b}\dot{p}_y$   & ${}^{b}\dot{p}_z$  & $\phi$ & $\theta$ & $\psi$ & $\omega_x$ & $\omega_y$ & $\omega_z$ \\ \hline\hline
Proposed                 & 0.66                & 0.05                & 0.24               & 4.93   & 2.31    & 0.32  & 0.12                & 0.04                & 0.02                \\
Baseline                 & 0.28                & 0.08                & 0.24               & 6.26   & 14.89   & 0.73  & 0.13                & 0.09                & 0.02                \\
\multicolumn{1}{l}{Unit} & \multicolumn{3}{c}{$\times 10^{-3} \mathrm{m^2/s^2}$} & \multicolumn{3}{c}{${}^\circ\!{}^2$} & \multicolumn{3}{c}{$\mathrm{rad^2/s^2}$}
\\ \hline
\end{tabular}
}
\label{Exp1MSE}
\end{table}

In Fig.~\ref{FTResultFigure}, the mean values of the proposed experiment are 0.04, -5.96, and -14.57 $\mathrm{N}$ and 1.58, -0.02, and 0.09 $\mathrm{N\cdot m}$ along the x-, y-, and z-axes, respectively, while those of the baseline are 0.2, -6.47, and -12.41 $\mathrm{N}$ and 1.62, -0.02, and 0.06~$\mathrm{N\cdot m}$. Overall, the differences between the proposed and baseline results are small relative to the signal variability and do not represent a practically significant deviation.

\begin{table}[!t]
\centering
\caption{Error under wrench disturbance}
\resizebox{\columnwidth}{!} {
\begin{tabular}{ccccccccccc}
\hline
                      &          & ${}^{b}\dot{p}_x$   & ${}^{b}\dot{p}_y$   & ${}^{b}\dot{p}_z$  & $\phi$ & $\theta$ & $\psi$ & $\omega_x$ & $\omega_y$ & $\omega_z$ \\ \hline\hline
\multirow{3}{*}{Mean} & Proposed & 0.20              & 0.82              & 0.46              & -0.82   & 9.51   & 1.15  & -0.73   & -0.08  & -0.50  \\
                      & Baseline & 0.12              & -0.61             & 1.33              & 0.99    & 9.09   & 2.06  & 0.15    & 0.32   & -0.16  \\
                      & Unit     & \multicolumn{3}{c}{$\times 10^{-2} \mathrm{m/s}$} & \multicolumn{3}{c}{${}^\circ$} & \multicolumn{3}{c}{$\times 10^{-2} \mathrm{rad/s}$} \\
\multirow{3}{*}{Std.} & Proposed & 0.07              & 0.08              & 0.06              & 1.94    & 2.44   & 1.39  & 0.48    & 0.44   & 0.26   \\
                      & Baseline & 0.08              & 0.11              & 0.07              & 2.52    & 3.44   & 1.84  & 0.51    & 0.54   & 0.29   \\
                      & Unit     & \multicolumn{3}{c}{$\mathrm{m/s}$} & \multicolumn{3}{c}{${}^\circ$} & \multicolumn{3}{c}{$\mathrm{rad/s}$} \\ \hline
\end{tabular}
}
\label{Exp2Error}
\end{table}

The tracking errors, computed as the differences between the desired and states, are illustrated in Fig.~\ref{Exp2ErrorFigure}, and the corresponding means and standard deviations are summarized in Table~\ref{Exp2Error}. Across ${}^{b}\dot{\mathbf{p}}$, $\mathbf{\Theta}$, and $\boldsymbol{\omega}$ the proposed approach reduces tracking-error oscillations, as indicated by consistently lower standard deviations on all axes. Relative to the baseline, the standard deviation decreases by 20.74\%, 23.36\%, 10.73\%, 23.21\%, 29.27\%, 24.83\%, 5.70\%, 19.51\% and 8.17\% respectively.

Under the wrench disturbance condition, the left columns of Table~\ref{Exp2Force} summarize the mean GRFs. The sums of the mean GRFs across legs for the proposed method and the baseline are 286.62~$\mathrm{N}$ and 288.32~$\mathrm{N}$, respectively. The mean GRFs of the front and hind legs are 122.97~$\mathrm{N}$ and 163.65~$\mathrm{N}$ for the proposed method, and 134.52~$\mathrm{N}$ and 153.8~$\mathrm{N}$ for the baseline, respectively. The difference between the front- and hind-leg GRFs is larger in the proposed method than in the baseline. In the right columns of Table~\ref{Exp2Force}, the force ratios are uniformly lower for the proposed method.

\begin{table}[!t]
\centering
\caption{GRF magnitudes and force ratios under wrench disturbance}
\small
\resizebox{\columnwidth}{!}{
\begin{tabular}{l*{4}{S[table-format=1.3]} c *{4}{S[table-format=1.3]}}
\toprule
& \multicolumn{4}{c}{GRF Magnitudes [N]} & & \multicolumn{4}{c}{Force Ratios} \\
& {RF} & {LF} & {RH} & {LH} && {RF} & {LF} & {RH} & {LH} \\
\hline\hline
Proposed & 68.23 & 54.74 & 89.21 & 74.44 && 0.115 & 0.179 & 0.126 & 0.146 \\
Baseline & 73.22 & 61.30 & 84.60 & 69.20 && 0.118 & 0.193 & 0.142 & 0.152 \\
\bottomrule
\end{tabular}}
\label{Exp2Force}
\end{table}

\begin{table}[t!]
\centering
\caption{MSE on grass terrain}
\resizebox{\columnwidth}{!} {
\begin{tabular}{cccccccccc}
\hline
                   & ${}^{b}\dot{p}_x$   & ${}^{b}\dot{p}_y$   & ${}^{b}\dot{p}_z$  & $\phi$ & $\theta$ & $\psi$ & $\omega_x$ & $\omega_y$ & $\omega_z$ \\ \hline\hline
Proposed           & 0.98              & 0.64              & 0.17              & 7.47    & 6.49    & 2.12  & 0.18    & 0.15   & 0.05   \\
Baseline           & 1.33              & 1.13              & 0.33              & 22.05   & 16.02   & 1.88  & 0.36    & 0.34   & 0.05   \\
Baseline{[}25Hz{]} & 0.90              & 0.51              & 0.18              & 4.22    & 4.21    & 1.22  & 0.21    & 0.14   & 0.05   \\
Unit               & \multicolumn{3}{c}{$\times 10^{-3} \mathrm{m^2/s^2}$}     & \multicolumn{3}{c}{${}^\circ\!{}^2$} & \multicolumn{3}{c}{$\mathrm{rad^2/s^2}$} \\ \hline
\end{tabular}
}
\label{Exp3MSE}
\end{table}

\subsection{Locomotion on Grass}

\begin{figure*}[t!]
    \centering
    \subfloat[\label{Exp3ControlVelFigure}]{
        \includegraphics[width=0.32\linewidth]{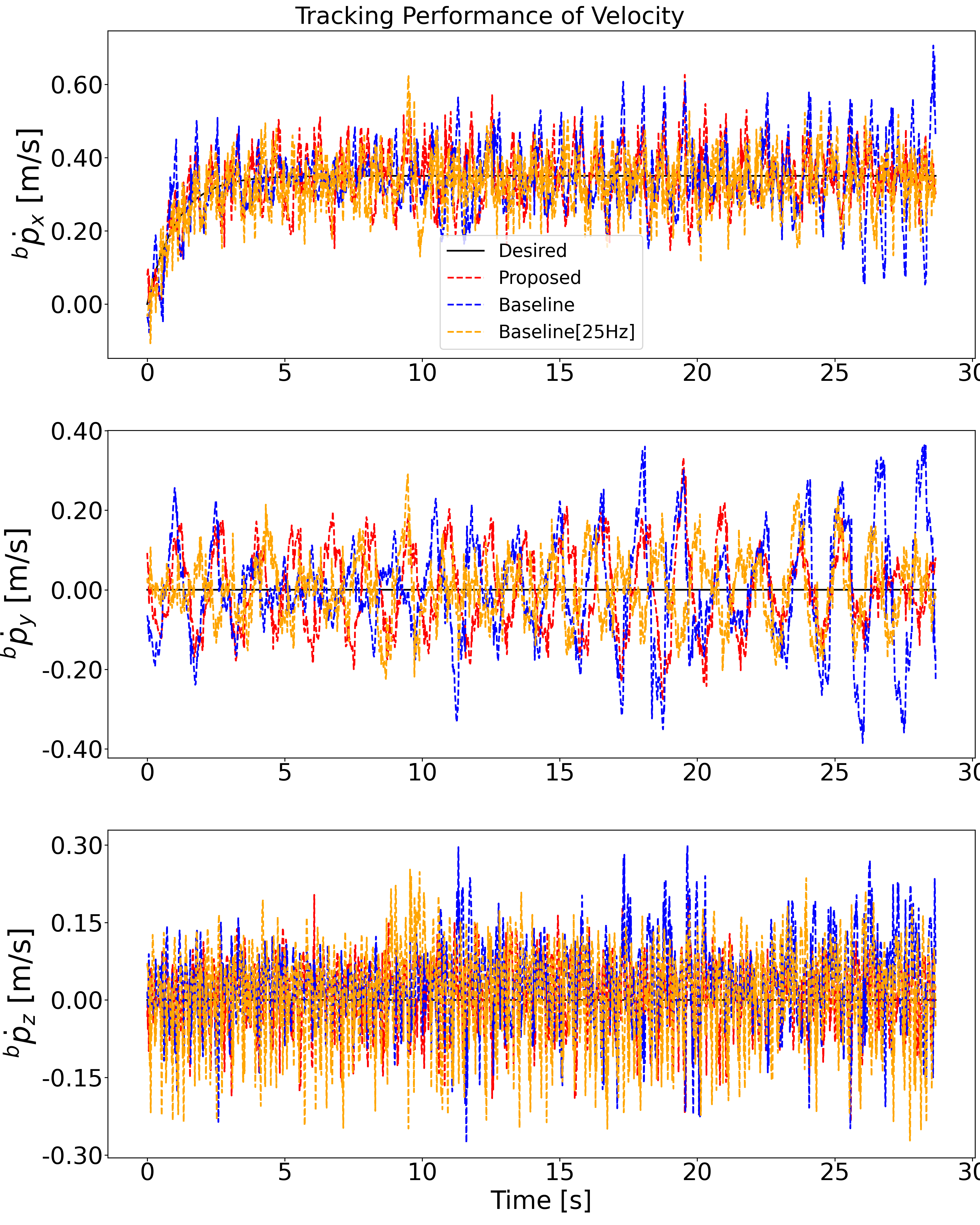}
    }%
    \subfloat[\label{Exp3ControlRPYFigure}]{
        \includegraphics[width=0.32\linewidth]{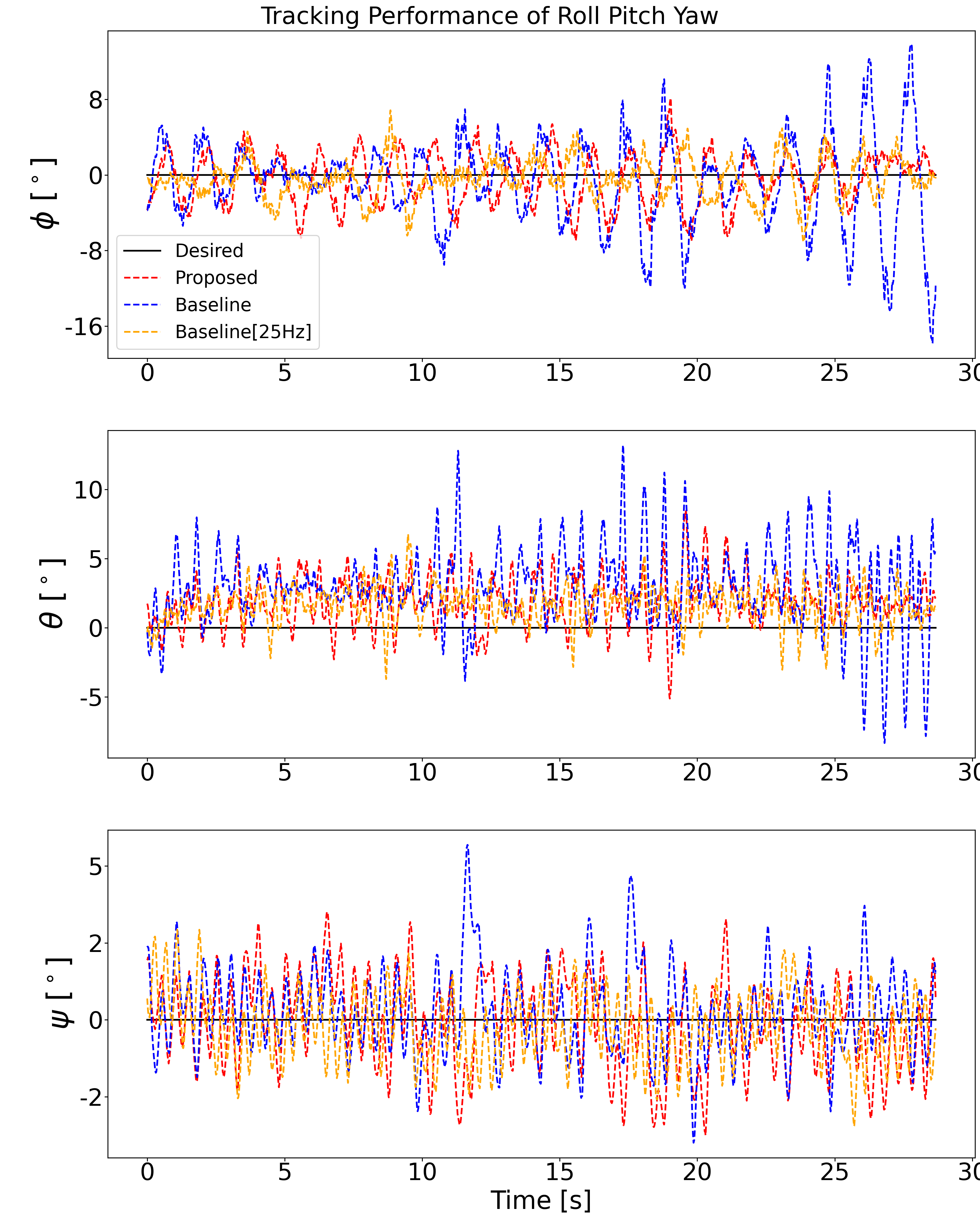}
    }%
    \subfloat[\label{Exp3ControlAngFigure}]{
        \includegraphics[width=0.32\linewidth]{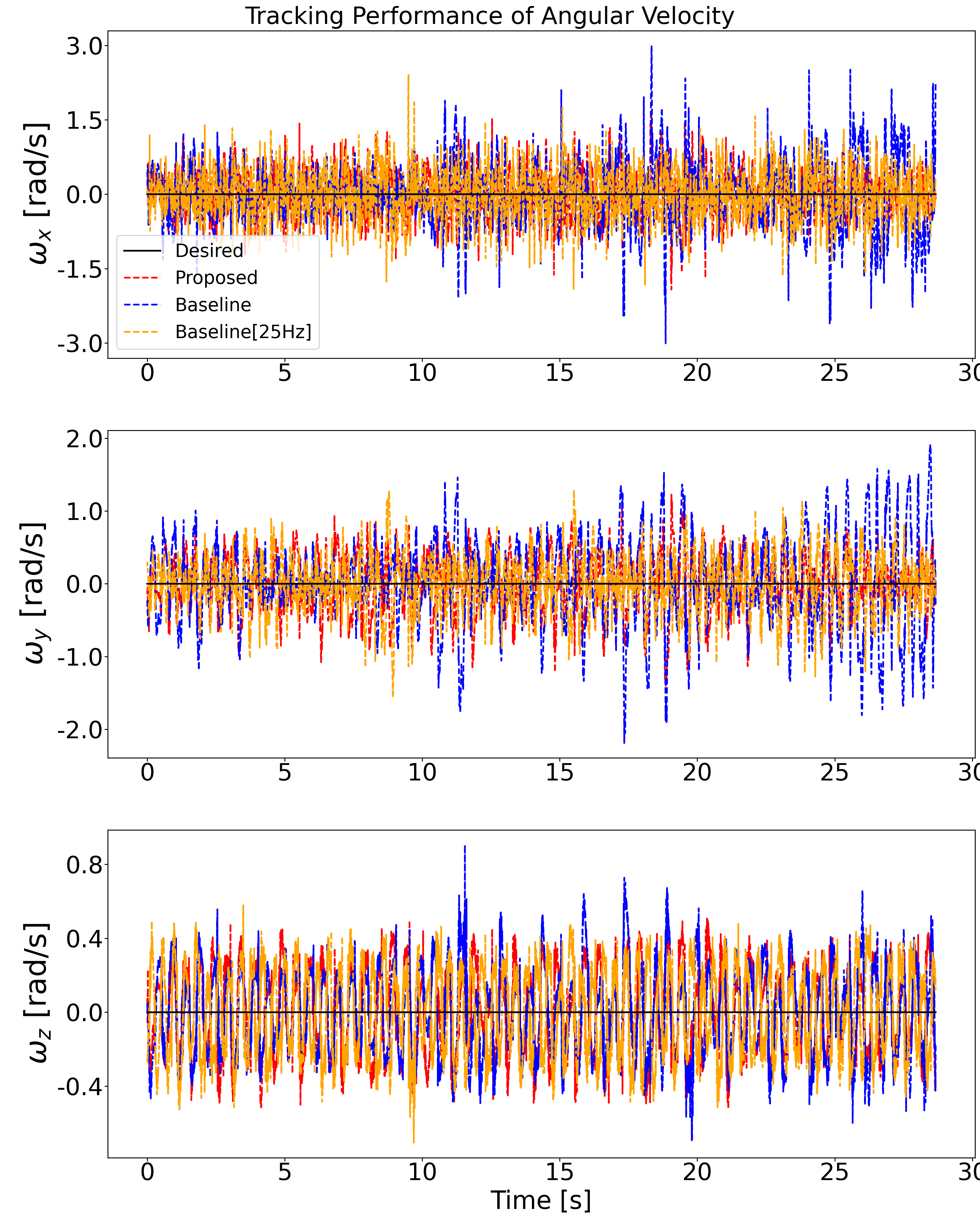}
    }%
    \caption{Experimental results of locomotion on grass terrain, illustrating desired values and states. (a)~Velocity in body frame. (b)~Roll, pitch, and yaw. (c)~Angular velocity.}
    \label{Exp3ControlFigure}
\end{figure*}

Unlike the rigid indoor floor, the grass terrain presents a deformable and uneven surface characterized by compliance and irregular height distribution. In this experiment, we additionally included a 25$\mathrm{Hz}$ baseline (200$\mathrm{ms}$ swing phase). The desired velocity along the body x-axis was specified as 0.35 $\mathrm{m/s}$.

The control performance of the three approaches is shown in Fig.~\ref{Exp3ControlFigure}, with the corresponding MSE values summarized in Table~\ref{Exp3MSE}. The proposed method achieves lower MSEs than the baseline for all states. In contrast, relative to the 25~$\mathrm{Hz}$ baseline, the MSEs increase in six states. In addition, as shown in Table~\ref{Exp3Force}, the sums of the mean GRFs across legs are 258.39, 281.62, and 344.20~$\mathrm{N}$ for the proposed, baseline, and 25~$\mathrm{Hz}$ baseline, respectively, indicating a monotonic increase from the proposed method to the baseline and then to the 25~$\mathrm{Hz}$ baseline. Across legs, the force ratios are lowest for the proposed method, followed by the 25~Hz baseline, and highest for the baseline.

\begin{table}[!t]
\centering
\caption{GRF magnitudes and force ratios on grass terrain}
\small
\resizebox{\columnwidth}{!}{
\begin{tabular}{l*{4}{S[table-format=1.3]} c *{4}{S[table-format=1.3]}}
\toprule
& \multicolumn{4}{c}{GRF Magnitudes [N]} & & \multicolumn{4}{c}{Force Ratios} \\
& {RF} & {LF} & {RH} & {LH} && {RF} & {LF} & {RH} & {LH} \\
\hline\hline
Proposed & 54.93 & 60.62 & 72.94 & 69.90 && 0.100 & 0.109 & 0.125 & 0.123 \\
Baseline & 69.70 & 76.87 & 69.95 & 65.10 && 0.141 & 0.138 & 0.163 & 0.170 \\
Baseline [25\,Hz] & 93.49 & 105.82 & 70.92 & 73.97 && 0.130 & 0.120 & 0.163 & 0.147 \\
\bottomrule
\end{tabular}}
\label{Exp3Force}
\end{table}

\begin{figure}[t!]
    \centering
        \includegraphics[width=0.99\columnwidth]{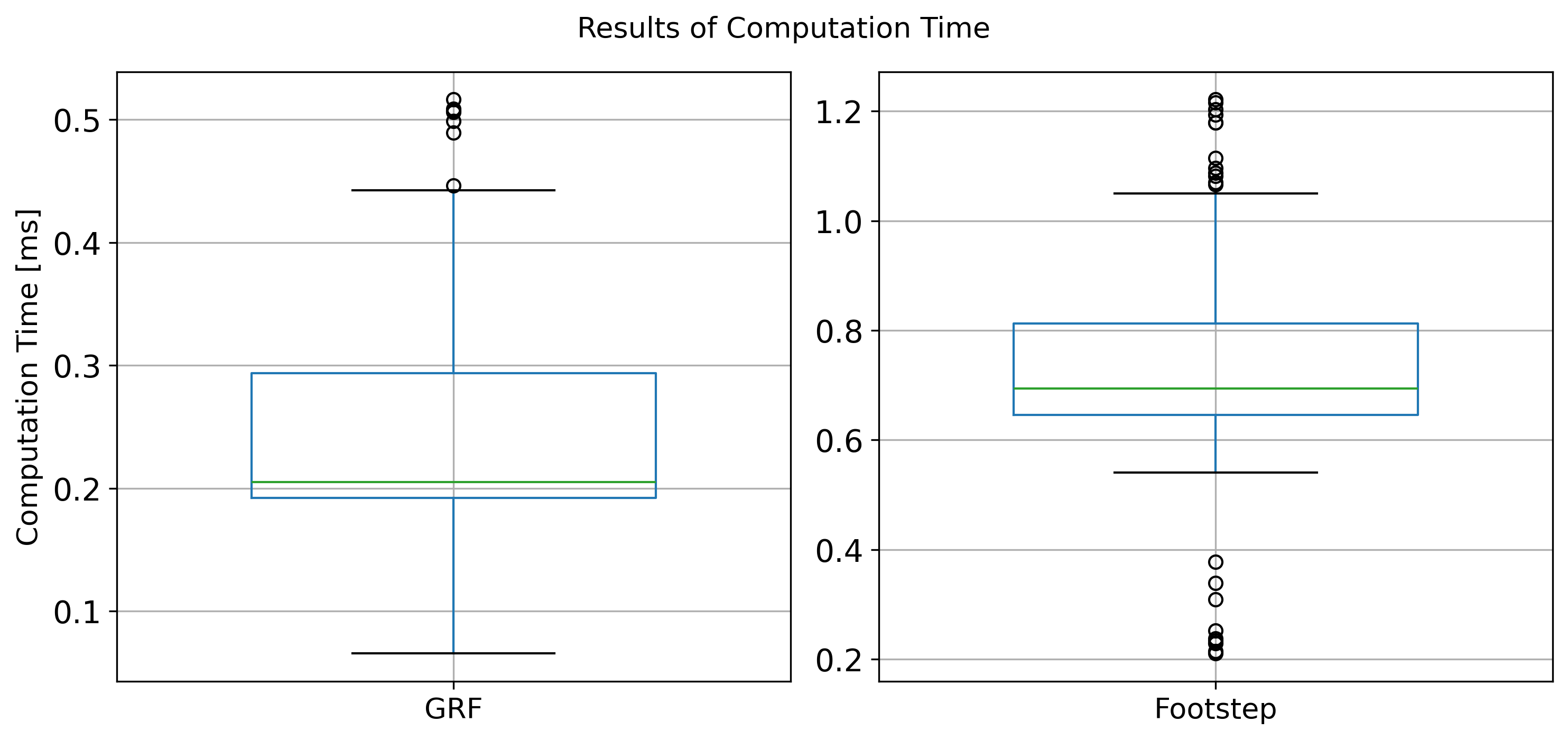}
        \caption{
        Boxplots of computation time during locomotion. Left~GRF~QP. Right~footstep~QP.
        }
    \label{computationtime}
\end{figure}

\subsection{Computation Time}
Fig.~\ref{computationtime} presents the computation time of the GRF and footstep MPCs during locomotion at 0.3~$\mathrm{m/s}$ along the x-axis in the body frame. The computation time was measured as the sum of the QP initialization and solve phases. The mean computation times of the GRF and footstep MPCs are 0.09~$\mathrm{ms}$ and 0.27~$\mathrm{ms}$, with standard deviations of 0.03~$\mathrm{ms}$ and 0.05~$\mathrm{ms}$, respectively. For the DMPC, the mean computation time is 0.36~$\mathrm{ms}$ with a standard deviation of 0.08~$\mathrm{ms}$, and the maximum observed value is 0.58~$\mathrm{ms}$.

\section{Discussion and Conclusion}
\label{sConclusion}
 In this paper, we studied the DMPC method for quadruped locomotion. This paper presented an MPC based footstep planning framework grounded in an approximated SRBM. The footstep MPC is hierarchically coupled with a GRF MPC through iterative feedback, so that both subproblems are updated simultaneously and progressively converge to GRF and footstep solutions that improve tracking performance. Extended stance and swing phase schedules are achieved through robust body-orientation regulation.

 The proposed method was experimentally validated on the Unitree GO1 platform. In the locomotion experiment on asymmetric-friction terrain, while the sum of mean GRF magnitudes remained comparable between methods, the proposed method reduced tracking errors in Euler angles and angular velocities and yielded lower force ratios across all legs. Under wrench disturbance, the total mean GRF magnitudes were similar between methods, whereas the proposed method reduced oscillations in all states and lowered force ratios for all legs. On grass, at the same update frequency, the proposed method reduced tracking errors across all states and exhibited lower total mean GRF magnitudes as well as lower force ratios. When the baseline update rate was increased to 25~$\mathrm{Hz}$, thereby shortening the swing phase, its tracking performance improved, but the total mean GRF magnitudes increased further. Applying the proposed method reduces the force ratio, which lowers the likelihood of slip in real-world operation and enables more efficient use of GRFs. Consequently, locomotion at higher velocities becomes feasible. Computation-time results indicate that the DMPC achieved real-time operation.

 The DMPC improved control performance and reduced the number of swing/stance phase transitions, thereby reducing impact loads at contact and improving hardware durability. Nevertheless, because the formulation relies on approximated dynamics, the resulting solutions are approximate. Future work will address this limitation by extending the framework to a nonlinear MPC that uses full-body dynamics, thereby providing a more rigorous validation of the proposed framework.

 \section*{Appendix A}
To prove that $\mathbf{P} \succeq0$, consider any $\mathbf{y} \in \mathbb{R}^{12j}, \mathbf{y}\neq\mathbf{0}$. By definition:

\begin{equation*}
    \mathbf{y}^{\top}\mathbf{Py} = 2\mathbf{y}^{\top}\mathbf{B}^{\top}_{qp}\mathbf{Q}\mathbf{B}_{qp}\mathbf{y} + 2\mathbf{y}^{\top}\mathbf{Ry} \,.
\end{equation*}
Since $\mathbf{R}$ is positive definite, it follows that:
\begin{equation*}
    2\mathbf{y}^{\top}\mathbf{Ry} > 0 \,.
\end{equation*}
For the first term:
\begin{equation*}
    2\mathbf{y}^{\top}\mathbf{B}^{\top}_{qp}\mathbf{Q}\mathbf{B}_{qp}\mathbf{y} = 2(\mathbf{B}_{qp}\mathbf{y})^{\top}\mathbf{Q}\mathbf{B}_{qp}\mathbf{y} \,.
\end{equation*}
Let $\mathbf{z} = \mathbf{B}_{qp}\mathbf{y}$. Substituting this into the equation gives:
\begin{equation*}
    2(\mathbf{B}_{qp}\mathbf{y})^{\top}\mathbf{Q}\mathbf{B}_{qp}\mathbf{y} = 2\mathbf{z}^{\top}\mathbf{Qz}\,.
\end{equation*}
Given that $\mathbf{Q}$ is positive definite, it follows that:
\begin{equation*}
    2\mathbf{z}^{\top}\mathbf{Qz} > 0 \,.
\end{equation*}
Combining these results:
\begin{equation*}
    \mathbf{y}^{\top}\mathbf{Py}>0, \forall\mathbf{y}\neq\mathbf{0} \,.
\end{equation*}
$\therefore \mathbf{P} \succ 0$. Hence, (\ref{qpProblem}) has a unique global minimum.

\vfill

\end{document}